\documentclass[journal,twoside,web]{ieeecolor}
\usepackage{generic}
\usepackage{cite}
\usepackage{amsmath,amssymb,amsfonts}
\usepackage{makecell}
\usepackage{threeparttable}
\usepackage{textcomp}
\usepackage{dsfont}
\usepackage[utf8]{inputenc} 
\usepackage[T1]{fontenc}    
\usepackage{hyperref}       
\usepackage{url}            
\usepackage{booktabs}       
\usepackage{amsfonts}       
\usepackage{nicefrac}       
\usepackage{microtype}      
\usepackage{graphicx}
\usepackage{xcolor}
\usepackage{colortbl,booktabs}
\usepackage{graphicx}
\usepackage{epstopdf}
\usepackage{blindtext, graphicx, mathrsfs, amsmath, amsfonts}
\usepackage{booktabs}
\usepackage{multirow, minipage-marginpar}
\usepackage{caption}
\usepackage{subfigure}
\usepackage{color,soul}
\usepackage[textsize=tiny]{todonotes}
\usepackage{xcolor}

\usepackage[linesnumbered,boxed,ruled,commentsnumbered]{algorithm2e}
\def\BibTeX{{\rm B\kern-.05em{\sc i\kern-.025em b}\kern-.08em
    T\kern-.1667em\lower.7ex\hbox{E}\kern-.125emX}}
\begin{document}
\title{Trust It or Not: Confidence-Guided Automatic Radiology Report Generation}

\author{Yixin Wang\textsuperscript{*}, Zihao Lin\textsuperscript{*}, Zhe Xu, Haoyu Dong, Jiang Tian, Jie Luo, Zhongchao Shi, Yang Zhang, \\Jianping Fan, and Zhiqiang He
\thanks{\textsuperscript{*} Y. Wang and Z. Lin contribute equally to this work. }
\thanks{Y. Wang and Z. He are with Institute of Computing Technology, Chinese Academy of Sciences, Beijing, China. Z. Lin and H. Dong are with Department of Electronic and Computer Engineering, Duke University, Durham, NC, USA. Z. Xu is with Department of Biomedical Engineering, The Chinese University of Hong Kong, Shatin, NT, Hong Kong, China. J. Luo is with Brigham and Women's Hospital, Harvard Medical School, Boston, MA, USA. J. Tian and Z. Shi are with AI Lab, Lenovo Research, Beijing, China. Y. Zhang is with Lenovo Corporate Research and Development, Lenovo Ltd., Beijing, China. J. Fan is with Department of Computer Science, University of North Carolina at Charlotte, Charlotte, NC, USA.}
\thanks{Corresponding authors: Y. Zhang (e-mail: zhangyang20@lenovo.com) and Z. He (e-mail: hezq@lenovo.com)}}

\newcommand{\suggest}[1]{\textcolor{red}{\{#1\}}}

\maketitle

\begin{abstract}
Medical imaging plays a pivotal role in diagnosis and treatment in clinical practice. Inspired by the significant progress in automatic image captioning, various deep learning (DL)-based methods have been proposed to generate radiology reports for medical images. Despite promising results, previous works overlook the uncertainties of their models and are thus unable to provide clinicians with the reliability/confidence of the generated radiology reports to assist their decision-making. In this paper, we propose a novel method to explicitly quantify both the visual uncertainty and the textual uncertainty for DL-based radiology report generation. Such multi-modal uncertainties can sufficiently capture the model confidence degree at both the report level and the sentence level, and thus they are further leveraged to weight the losses for more comprehensive model optimization. Experimental results have demonstrated that the proposed method for model uncertainty characterization and estimation can produce more reliable confidence scores for radiology report generation, and the modified loss function, which takes into account the uncertainties, leads to better model performance on two public radiology report datasets. In addition, the quality of the automatically generated reports was manually evaluated by human raters and the results also indicate that the proposed uncertainties can reflect the variance of clinical diagnosis.
\end{abstract}

\section{Introduction}
Radiology report generation, which combines Computer Vision (CV) and Natural Language Processing (NLP) to automatically generate reports for medical images, has become a rapidly-growing topic in the medical imaging field. Since generating hand-crafted diagnostic reports is time-consuming and hard to follow the same protocol, automatic generation can alleviate doctor’s workload and make standardized diagnosis. With the convincing performance and understandable interpretations of DL's predictions, there is a great potential to automate the clinical workflows. Most existing research for radiology report generation usually borrows ideas from natural image captioning   \cite{COATT,report02,report03,report04,report06,report07}, where the images are firstly encoded into vectorial representations and then decoded to generate the captions. Despite promising results have been obtained by these DL methods, their reliability has not been explored sufficiently for the following reasons.
First, learning from inadequate samples can result in unreliable models, which is prone to generate inaccurate reports. Second, existing evaluation metrics are mostly designed for mainstream NLP tasks, which have been proved to be not suitable for the task of generating radiology reports from medical images \cite{MONSHI2020101878, Re-evaluating}. 
However, it has been extensively shown in \cite{reportuncer01, reportuncer02} that some traditional factors including poor image quality, improper protocol selection, deficient clinical data, limitations in experience and technology, and lack of established diagnostic standards may affect diagnosis report uncertainty and thus interfere the clinical outcomes. Reiner BI \cite{reportuncer03} also indicated the need for report uncertainty in clinical workflow and built their relationships. As mentioned in \cite{reportuncer03}, the referring clinician may require follow-up radiographs due to the inherent uncertainty contained in the report. If, however, the clinician was presented with context and user-specific uncertainty data, they could make a further decision on the need for follow-up radiographs. Similarly, as the DL-based automatic report generation systems are data-driven and less interpretable, they can also introduce more uncertainties to the predicted reports and thus become less reliable to clincians.
Therefore, it is highly desirable to assess how much patients and doctors can trust such generated radiology reports for decision making. This motivates us to present a comprehensive confidence measurement to quantify the model uncertainty in radiology report generation. 

Most existing research on model uncertainty characterization focused on single-modal tasks. For example, \cite{uncer01,uncer02,classification01,huang2018efficient,uncer03} studied the model uncertainties for the classification/segmentation tasks, where additional supervisions are leveraged to promote the model's prediction reliability. \cite{xiao2019quantifying, xu2020understanding} quantified the uncertainties for different NLP tasks, such as sentiment analysis, named entity recognition, language modeling and summarization. For multi-modal tasks such as image captioning, however, estimating its uncertainties is challenging because the uncertainties may appear in the processes of both visual feature extraction and text generation. 

In this paper, we propose a new method to characterize and quantify both the visual and textual uncertainties in radiology report generation. For the visual encoder, we first adopt Monte Carlo (MC) dropout \cite{MCdropout} to sample weights as a Bayesian Neural Network (BNN). Since it is unrealistic to directly quantify the model uncertainty for visual extraction in the output space, we then design an extra AutoEncoder to obtain such visual uncertainty, \textbf{VISVar}, which can estimate the parameter variance for the visual extractor and encourage the model to learn meaningful and generalizable latent representations. Besides, for the generated radiology reports, a proper report similarity measurement formulates the basis of uncertianty estimation. Inspired by Word Rotator’s Distance (WRD) \cite{yokoi2020word}, we propose the \textbf{S}entence \textbf{M}atched \textbf{A}djusted Semantic \textbf{S}imilarity (SMAS) to measure their similarities, which considers sufficient information and characteristics of medical diagnosis reports. Based on SMAS, \textbf{SMASVar} and $\textbf{SMASVar-}\boldsymbol{l}$ are further proposed to separately measure the model uncertainties at both the report level and the sentence level. Particularly, both uncertainty measurements focus on the semantics of sentences rather than their structures, which can provide more valuable confidence scores for the generated radiology reports. Finally, such visual-textual uncertainties are leveraged to define an uncertainty-weighted loss to achieve more comprehensive model optimization, lessen misjudgement risk and improve overall performance on radiology report generation. Our extensive experiments conducted on the Indiana University chest X-Ray (IU X-Ray) and a public COVID-19 CT Report dataset (COV-CTR)\cite{li2020auxiliary} have demonstrated the effectiveness of our proposed method. More importantly, our method is better at characterizing the visual-textual uncertainties and providing more reliable confidence scores for radiology report generation than the state-of-the-art works do.

Our contributions are as follows.
\begin{itemize}
\item We are the first to explicitly quantify the inherent visual-textual uncertainties existing in the multi-modal task of radiology report generation from the perspective of both report level and sentence level. 
\item We demonstrate the significance of visual extractor in image-caption architectures and incorporate an auxiliary AutoEncoder branch to encode specific features and further obtain visual uncertainty.
\item We propose a novel method SMAS to measure the semantic similarity of radiology reports, which can better capture the characteristics of diagnosis information. Textual uncertainty is further quantified based on SMAS.
\item The obtained uncertainties can be integrated into the model to balance the loss function of reports/sentences and benefit the whole optimization process.
\item Extensive experiments evaluated by both commonly used automatic metric and experienced doctors show that our approach leads to more accurate generated radiology reports and our uncertainty estimation can better capture the relative confidence among different reports and sentences, which yields valuable guidance for report generation in clinical practice.
\end{itemize}
\section{Related Work}
\begin{figure*}[t]
\includegraphics[width=2\columnwidth]{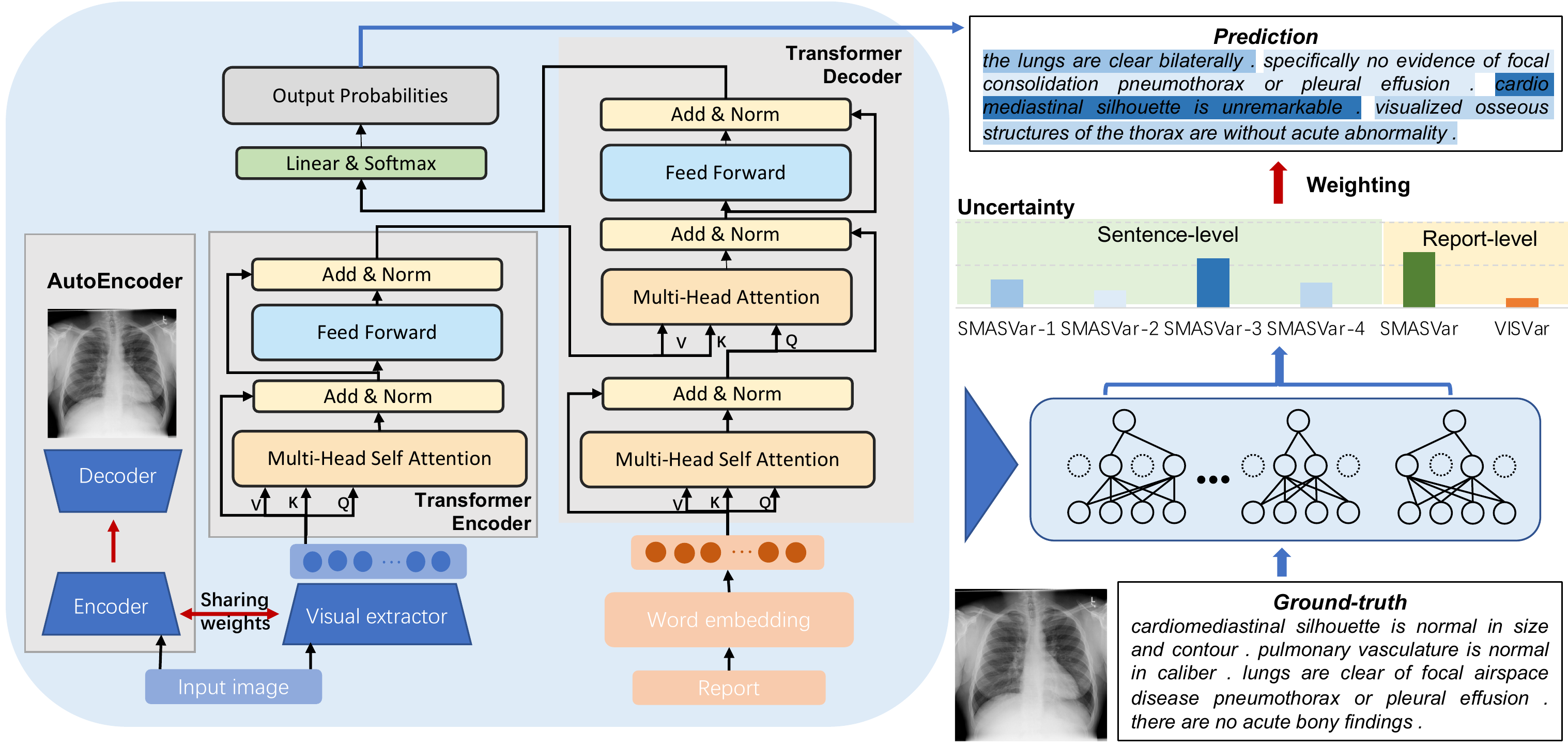}
\caption{Illustration of the proposed method. The left part presents the architecture of the image encoder, which consists of an AutoEncoder for image reconstruction, a transformer encoder to extract features and a transformer decoder to predict words via attention mechanism. The right part presents the uncertainty estimation workflow. }
\label{architecture}
\end{figure*}
Our work has drawn recent attention on 1) Radiology Report Generation and 2) Uncertainty Modeling.
\subsubsection{Radiology Report Generation}
For a given medical image, radiology report generation aims at automatically generating a set of descriptive and informative sentences. It is related to the computer vision task for image captioning, where various CNN-RNN/LSTM architectures together with attention mechanisms have achieved promising results \cite{imgcap01,imgcap02,imgcap03,imgcap04,jbhi01,jbhi02}. Recently, inspired by the capacity on parallel training, transformers \cite{imgcap05,imgcap06,imgcap07,imgcap08} have been successfully applied to predicting words according to the extracted features from CNN. Generating radiology reports is much more challenging as it requires to identify the relevant information from noisy radiology images with similar patterns and generating long contextual descriptions \cite{COATT,report02,report03,report04,report06,report07,report08,report09,report10}. Particularly, Liu et al. \cite{pmlr-v106-liu19a} proposed a domain-aware automatic generation system, which employed a hierarchical convolutional-recurrent neural network and reinforcement learning for report generation. Chen et al. \cite{report07} incorporated the relational memory into the transformer via a relational memory and memory-driven conditional layer normalization, which is capable of generating long reports with informative content. Liu et al. \cite{report08} explored the posterior and prior knowledge from a pre-defined medical knowledge graph to alleviate visual and textual data bias. Wang et al. \cite{report09} proposed an auxiliary image-text matching task to learn strongly correlated visual and text features and then the two tasks could boost each other in a progressive way. Chen et al. \cite{report10} designed a shared memory to record the alignment between images and texts, which could facilitate the interaction and generation across modalities. However, these methods only focus on achieving higher generation accuracy, but they do not actually evaluate whether the generated reports are applicable in real clinical practice.
\subsubsection{Uncertainty Modeling}
Uncertainty mostly occurs due to the inadequate knowledge and data during model training, and uncertainty estimation is a key approach to quantify the reliability of the model's predictions\cite{uncer02}. BNN interprets a posterior uncertainty of the underlying model parameters, which can be realized via sampling weights and forwarding the inputs through the networks for multiple times. Recently, there has been increasing number of studies on investigating the uncertainty in different CV and NLP tasks. These methods mostly quantify the uncertainties through the predicted classes such as segmentation \cite{huang2018efficient,uncer03,uncer04}, detection \cite{detection01,detection02}, classification \cite{classification01,classification02}, sentiment analysis and named entity recognition \cite{xiao2019quantifying}. Machine translation or other text generation tasks usually quantify the uncertainties through Beam Score, Sequence Probability \cite{wang2019improving} or BLEU Score \cite{xiao2020wat}.

Compared with the above tasks, quantifying uncertainties in radiology report generation task is more challenging because it is hard to estimate visual uncertainty with only feature extractors and to quantify the textual uncertainty of key information in diagnostic reports.
Therefore, we first design an AutoEncoder as a BNN to quantify visual uncertainty in the output level. To get textual uncertainty, we need to quantify the semantic similarity between two diagnostic reports. Simply calculating similarity from commonly used scores such as BLEU \cite{xiao2020wat}, ROUGE \cite{lin2004rouge} and METEOR \cite{banerjee2005meteor}, or image captioning metrics such as CIDEr \cite{vedantam2015cider} and SPICE \cite{anderson2016spice} fail to capture semantic information.
BERT-based pre-trained methods such as BERTScore \cite{zhang2019bertscore} or Sentence-BERT \cite{reimers2019sentence} have been shown to be ineffective for unsupervised semantic textual similarity (STS) task \cite{yokoi2020word}. Recent transformer-based methods such as SIF \cite{chang2021extending} are unable to capture medical knowledge without costly training. Word Rotator's Distance (WRD) \cite{yokoi2020word} performs better on STS task and it can capture medical information with proper pre-trained medical word embeddings. However, measuring similarity on diagnostic report is different from general semantic similarity measurement, because: (1) it is inappropriate to treat diagnostic reports as single long sentences, as sentences from the same report can be highly independent; (2) some key words or descriptions of diseases will greatly affect the semantic similarity between two reports. To address these problems, we consider WRD at the sentence-level to measure the semantic similarity between radiology reports. This leads to our novel diagnostic report similarity measurement, \textbf{S}entence \textbf{M}atched \textbf{A}djusted Semantic \textbf{S}imilarity (SMAS).

It is a promising and under-explored direction as the medical-specific uncertainty can provide information on how confident the generated reports are and in turn enables time-effective corrections by AI diagnosis itself and doctors.

\section{Method}
Fig. \ref{architecture} shows an overview of our proposed method. We first illustrate the image-text generation network in Section \ref{sec-3.2}-\ref{sec-3.4}. Section \ref{sec-3.5} describes how to estimate the report-level and sentence-level uncertainties in the proposed architecture. We elaborate our novel uncertainty-weighted loss in Section \ref{sec-3.6}.
\subsection{Feature AutoEncoder}
\label{sec-3.2}
The raw images are taken as the input $\mathbf{\{X\}_{i=1}^{N}} \in \mathcal{R}^{3 \times 224 \times 224}$ for the image encoder, which aims at extracting high-dimensional visual features to detect a wide range of visual concepts, landmarks and other entities for the language model. The image encoder adopts a pre-trained Convolution Neural Network (CNN) from ImageNet \cite{imagenet}, and the parameters are fixed during training. However, medical images, being visually different from natural images, contain more specific physiological and medical-related information. Since text generation largely relies on the content of the input images, the extracted representations should be some specific medical features for a final description. 

For the above purposes, we design an auxiliary AutoEncoder branch to add an additional guidance to the feature extractor for image reconstruction. The auxiliary module allows the image encoder to extract features that are more effective in addressing the target-specific problem. 
It also introduces additional regularization to the framework, which alleviates the data deficiency issue.
Specifically, a shared encoder, which is pre-trained on ResNet-101, maps $\mathbf{X}$ to hidden representations $\mathbf{R} \in \mathcal{R}^{2048 \times 7 \times 7}$. To preserve more spatial content, the feature maps from the intermediate layer are picked for the decoder part of AutoEncoder, with the size of $\mathbf{Y} \in \mathcal{R}^{512 \times 28 \times 28}$. We denote the mapping function $f(\cdot)$ as
\begin{equation}
    \mathbf{Y}=f(\mathbf{X})=s_{f}\left(W \mathbf{X}+b_{\mathbf{X}}\right),
\end{equation}
where $s_{f}$ is a nonlinear activation function and $W$ and $b$ represent the weight matrix and bias.
The decoder part of AutoEncoder is trained to reconstruct $Y$ back to $X'$ via a mapping $g(\cdot)$ as
\begin{equation}
     \mathbf{X'}=g(\mathbf{Y})=s_{g}\left(W' \mathbf{Y}+b_{\mathbf{Y}}\right).
\end{equation}
The whole process is performed in an unsupervised manner which empowers the encoder with strong ability on automatic feature extraction. The reconstruction loss is generally a mean squared error (MSE) as
\begin{equation}
    \ell_{AutoEn}(\theta)=\sum_{i=1}^{N}\left\|\mathbf{X_{i}}-\mathbf{X_{i}^{\prime}}\right\|^{2}=\sum_{i=1}^{N}\left\|\mathbf{X_{i}}-g\left(f\left(\mathbf{X_{i}}\right)\right)\right\|^{2}.
\end{equation}

\subsection{Transformer Encoder}
\label{sec-3.3}
The obtained abundant task-specific features $\mathbf{R}$ are then used as $N$ input tokens with $d=2048$ dimensions to a standard transformer encoder. Each encoder layer consists of a multi-head self-attention layer followed by a positional feed-forward neural network. Specifically, the tokens are first transformed into query $Q \in \mathcal{R}^{N_{q} \times d_{k}}$, key $K \in \mathcal{R}^{N_{k} \times d_{k}}$ and value $V \in \mathcal{R}^{N_{k} \times d_{v}}$ as
\begin{equation}
    Q=\mathbf{R} W_{Q}, \quad K=\mathbf{R} W_{K},\quad V=\mathbf{R} W_{V},
\end{equation}
where $W_{Q}, W_{K}, W_{V}$ are the learnable projection matrices. Then the scaled dot-product attention is applied by each attention head $\textit{head}_{i}$ as
\begin{equation}
    \textit{head}_{i} =\operatorname{Attention}(Q, K, V)=\operatorname{Softmax}\left(\frac{Q K^{T}}{\sqrt{D_{k}}}\right) V.
\end{equation}
All the results of different heads are aggregated by a linear transformation $W_{O}$ as
\begin{equation}
    \operatorname{MultiHead}(Q, K, V)=\operatorname{Concat}\left(\textit{head}_{1}, \ldots, \textit { head}_{h}\right) W_{O}.
\end{equation}
Besides, the residual connections and layer normalization $\operatorname{Norm}$ are performed, and the feed-forward operation $\operatorname{FFN}$ is adopted in the end of each block. The output of the encoder is formalized as
\begin{equation}
\begin{split}
 \mathbf{R_{m}}&=\operatorname{Norm}(\mathbf{R}+\operatorname{MultiHead}(Q, K, V)),\\
 \mathbf{R^{\prime}}&=\operatorname{Norm}\left(\mathbf{R_{m}}+ \operatorname{FFN} \left(\mathbf{R_{m}}\right)\right).
\end{split}
\end{equation}
\subsection{Transformer Decoder}
\label{sec-3.4}
The transformer decoder consists of a multi-head self-attention module and a multi-head cross-attention module. The former receives the embedding matrix of the current report and outputs the tokens $\mathbf{h_{a}} = \{ \mathbf{h_{1}, h_{2},..., h_{A}}\}$ with length $A$. The output $\mathbf{R^{\prime}}$ of the encoder is utilized as the key and value to the multi-head cross-attention, with $\mathbf{h_{a}}$ treated as the query. It can be formalized as
\begin{equation}
    \mathbf{h_{m}}=\operatorname{Norm}(\mathbf{h_{a}}+\operatorname{MultiHead}(\mathbf{h_a}W_{h}, \mathbf{R^{\prime}}W_{R'}, \mathbf{R^{\prime}}W_{R'})).
\end{equation}
Similar to the encoder, $\mathbf{h_{m}}$ is processed by $\operatorname{FFN}$ and $\operatorname{Norm}$ as
\begin{equation}
    \mathbf{h^{\prime}}=\operatorname{Norm}\left(\mathbf{h_{m}}+ \operatorname{FFN} \left(\mathbf{h_{m}}\right)\right).
\end{equation}
The context vector is then fed into a linear layer and the output is used to predict the probability of words. 
\subsection{Uncertainty Estimation}
\label{sec-3.5}
The generated reports can be successfully derived from the aforementioned model. However, there is still no estimation about the reliability level of such predictions, which is a crucial reference for doctors to make further decisions. To tackle this problem, we propose a new method to evaluate the uncertainties of the generation system, which are defined as \textbf{vis}ual uncertainty measured by \textbf{var}iance \textbf{VISVar} and textual \textbf{SMAS} uncertainty measured by \textbf{var}iance \textbf{SMASVar}. These two uncertainties quantify the model's uncertainty from the perspective of the processes of image embedding and word generation, respectively. To estimate this uncertainty, we adopt Monte Carlo (MC) dropout variational inference method \cite{MCdropout}. Particularly, any neural network with dropout can be viewed as a BNN which interprets the probability distribution over the model's parameters. With the dropout activated, the model uncertainties can be estimated by performing $T$ stochastic forward passes on the model, which samples from the approximate posterior.
\subsubsection{Visual Uncertainty}
Since it could be difficult to perform uncertainty quantification for the feature extractor in the output space, our proposed AutoEncoder branch serves as a guidance to obtain the uncertainty map for the input image. This model is extended into a BNN, which captures uncertainty through Bayesian posterior of the reconstruction regression task. The posterior predictive distribution $p(\mathbf{X'} \mid \mathbf{X})$ of $\theta$ can be appropriated as:
\begin{equation}
    p(\mathbf{X'} \mid \mathbf{X})=\int p(\mathbf{X'} \mid \boldsymbol{\theta}, \mathbf{X}) p(\boldsymbol{\theta} \mid \mathbf{X}) \mathrm{d} \boldsymbol{\theta}.
\end{equation}
As this is intractable in deep networks, we apply dropout variational inference method \cite{MCdropout} to acquire the approximation of the true posterior distribution. In particular, the dropout operation is utilized after each non-linear layer of the shared part of the AutoEncoder as we expect to obtain the posterior distribution of the feature extractor in our report generation framework. Given an input, the dropout is enabled and the output predictive distributions are obtained through $T$ times stochastic forwarding pass the network. By averaging $T$ samples, the mean $\boldsymbol{\mu}_{\text{VIS}}'$ and the variance $\boldsymbol{\sigma'}_{\text{VIS}}^{2}$ are obtained as
\begin{equation}
\begin{split}
    & \boldsymbol{\mu}_{\text{VIS}}'=\mathbf{\overline{X'}}=\frac{1}{T} \sum_{t=1}^{T} \mathbf{{X'}_{t}},\quad \\
 & \boldsymbol{\sigma'}_{\text{VIS}}^{2}=\frac{1}{T} \sum_{t=1}^{T}\left(\mathbf{{X'}_{t}}-\frac{1}{T} \sum_{t=1}^{T} \mathbf{{X'}_{t}}\right)^{2}.
\end{split}
\end{equation}
In this way, $\boldsymbol{\sigma'}_{\text{VIS}}^{2}$ is obtained as an uncertainty map for the corresponding input. The overall visual uncertainty is defined as the mean over all pixels of $\mathbf{X}$, i.e., $\textbf{VISVar} = \overline{\left|\boldsymbol{\sigma'}_{\text{VIS}}\right|}$, which represents the report-level visual uncertainty value.
\subsubsection{Textual Uncertainty} \label{Generative Uncertainty}

The textual uncertainty can be obtained by the variance of semantic similarity scores among generated diagnostic reports. Combined with medical word embeddings (such as BioWordVec \cite{zhang2019biowordvec}), WRD \cite{yokoi2020word} can be a feasible similarity measurement for reports. Given that sentences from the same radiology report are highly independent of each other, we propose a new method called Sentence-Matched Adjusted Similarity (SMAS), which derives from WRD \cite{yokoi2020word}. Based on SMAS, we further propose \textbf{SMASVar} to measure the report-level textual uncertainty and $\textbf{SMASVar-}\boldsymbol{l}$ to measure the sentence-level textual uncertainty.

WRD is based on Earth Mover's Distance (EMD) (also called Wasserstein distance) \cite{hormander2006grundlehren, santambrogio2015optimal}. Intuitively, EMD measures the minimum energy cost that is needed to turn one pile of dirt to another. In WRD, a sequence of word vectors, which are usually gained from word embeddings pre-trained by Word2Vec \cite{mikolov2013efficient} or FastText \cite{bojanowski2017enriching}, serves as the pile of dirt and the energy consumption is the transfer distance from one sentence to another per unit information transfer.

Assume that the sentence $S$ has word vectors $\boldsymbol{w}_1, \boldsymbol{w}_2, ..., \boldsymbol{w}_n$, the sentence $S'$ has word vectors $\boldsymbol{w'}_1, \boldsymbol{w'}_2, ..., \boldsymbol{w'}_m$, and the cosine similarity $d_{i, j}$ represents the distance between $\boldsymbol{w}_i$ and $\boldsymbol{w'}_j$. We denote the optimal solution as $\lambda_{i, j}$ which represents turning $\lambda_{i, j}$ amount of information from $\boldsymbol{w}_i$ to $\boldsymbol{w'}_j$. Thus, measuring the similarity between two sentences $S$ and $S'$ becomes an optimal problem and can be expressed as:
\begin{equation}
\begin{split}
 \operatorname{WRD}(S,S') & = \min _{\lambda_{i, j} \geq 0} \sum_{i, j} \lambda_{i, j} d_{i, j} \\
 \text {s.t.} \quad \sum_{j} \lambda_{i, j}&=p_{i}, \quad \sum_{i} \lambda_{i, j}=q_{j}, \\
 d_{i, j}&=1-\frac{\boldsymbol{w}_{i} \cdot \boldsymbol{w'}_{j}}{\left\|\boldsymbol{w}_{i}\right\| \times\left\|\boldsymbol{w'}_{j}\right\|},
\end{split}
\end{equation}
where $p_i$ and $q_j$ represent the information contained in the words $\boldsymbol{w}_{i}$ and $\boldsymbol{w'}_{j}$.
Moreover, according to \cite{yokoi2020word}, the information contained in each sentence ${I_S}, {I_{S'}}$ is obtained through summing the norm of each word vector:
\begin{equation}
\begin{split}
    &p_{i}=\frac{\left\|\boldsymbol{w}_{i}\right\|}{I_S}, \quad I_S=\sum_{i=1}^{n}\left\|\boldsymbol{w}_{i}\right\|; \\ &q_{j}=\frac{\left\|\boldsymbol{w'}_{j}\right\|}{I_{S'}}, \quad I_{S'}=\sum_{j=1}^{m}\left\|\boldsymbol{w'}_{j}\right\|.
\end{split}
\end{equation}
 
WRD describes the distance between two sentences, and we denote Word Rotator’s Similarity $(\text{WRS}) = 1-\text{WRD}$ as the sentence similarity between them. Given two reports $R=(S_{1}, S_{2}, ..., S_{L})$ and $R'=(S'_{1}, S'_{2}, ..., S'_{L'})$, where $L$ and $L'$ represent the numbers of sentences, the similarity between $R$ and $R'$ can be obtained by averaging the WRS of corresponding sentences in the two reports. We design a $\operatorname{MATCH}(\cdot)$ algorithm to find their corresponding sentences. Firstly, we get a dict of similarity scores $L * L'$ of each pairs of sentences $\{S_{i}, S'_{j}\}$, where $i$ and $j$ are any of the index in $R$ and $R'$. We then sort the similarity scores in descending order and pop the top item out iteratively. After each iteration, items containing the same sentences will be deleted until there is no item left in the dict. The selected sentences pairs are the matched sentences of the two reports. To speed up the matching process, we do not use WRS to get the similarities of sentences. Instead, we obtain the sentence embeddings through BioSentVec \cite{chen2019biosentvec} operation $\operatorname{BioV}(\cdot)$ and then calculate the cosine similarities $\operatorname{COS}(\cdot)$ of each sentence pairs.
All matched sentences pairs $\{S_{i}, S'_{j}\}$ of report $R$ and $R'$ can be obtained by
\begin{equation}
\{S_{i}, S'_{j}\} = \operatorname{MATCH}_{i, j} \left(\operatorname{COS}(\operatorname{BioV}\left(S_i\right), \operatorname{BioV}\left(S'_j)\right)\right).
\end{equation}

The sentence matched similarity (SMS) between reports $R$ and $R'$ is calculated by
\begin{equation}
\label{WRS}
\operatorname{SMS}(R, R') =\frac{2}{L + L'} \sum_{\{S_{i}, S'_{j}\}}\operatorname{WRS}(S_{i}, S'_{j}).
\end{equation}

Based on the characteristics of radiology reports, we assume that each sentence conveys independent information in a given report. Thus, the inconsistency between $L$ and $L'$ reflects the redundant or missing information (also explained in Section \ref{smas-exp}). To eliminate this negative effect, we multiply a penalty item to SMS to get Sentence-Matched Adjusted Similarity (SMAS) of report $R$ and $R'$:

\begin{equation}
\operatorname{SMAS}(R, R') = \operatorname{SMS}(R, R') * \left(1 - \frac{|L - L'|}{\max(L, L')}\right).
\end{equation}

With $T$ samples generated by $T$ times MC Dropout, the variance of each report can be calculated by
\begin{equation}
\begin{aligned}
\hat{\boldsymbol{\sigma}}_{\text{SMAS}}^{2}  \approx\!\frac{1}{(T-1)!}\!&\sum_{i=1}^{T}\! \sum_{j=i+1}^{T}\!\Bigg(\!\operatorname{SMAS}\left(R_{\theta_i}, R_{\theta_j}\right)\! \\ 
& - \!\frac{1}{(T-1)!} \sum_{p=1}^{T}\!\sum_{q=p+1}^{T}\!\!\operatorname{SMAS}\!\left(\!R_{\theta_p}, \!R_{\theta_q}\!\right)\!\!\Bigg)^{2}
\end{aligned}
\end{equation}
where $R_{\theta_i}$ and $R_{\theta_j}$ denote the outputs of the models with different parameters $\theta_i$ and $\theta_j$.
The report-level textual uncertainty for each report, $\textbf{SMASVar} = \left|\hat{\boldsymbol{\sigma}}_{\text{SMAS}}\right|$, along with \textbf{VISVar}, represents the report-level uncertainty. To obtain the sentence-level uncertainty, we first find an average reference report among all $T$ samples. The average report has the smallest distance with the rest of $T-1$ samples \cite{xiao2020wat}. Therefore, the reference report $\tilde{R}$ can be obtained by
\begin{equation}
\tilde{R}=\underset{R_{\theta_i}}{\arg \min }\left(\sum_{\forall j \neq i}^{T}\left(1-\operatorname{SMAS}\left(R_{\theta_i}, R_{\theta_j}\right)\right)\right) .
\end{equation}
Denoting $\tilde{S}_{l}$ as the $l^{th}$ sentence in $\tilde{R}$, and $S_{\theta_j, l}$ as the corresponding sentence in report $R_{\theta_j}$ with $\tilde{S}_{l}$, the sentence-level uncertainty of the $l^{th}$ sentence in the reference report is defined as $\textbf{SMASVar-}\boldsymbol{l} = \left|\hat{\boldsymbol{\sigma}}_{\text{SMAS-}l}\right|$, where
\begin{equation}
\begin{aligned} 
\hat{\boldsymbol{\sigma}}_{\mathrm{SMAS}-l}^{2} \approx \frac{1}{T-1} &\sum_{j \neq \mathrm{ref}}^{T}\Bigg(\operatorname{WRS}\left(\tilde{S}_{l}, S_{\theta_{j}, l}\right) \\ 
&-\frac{1}{T-1}\!\sum_{p \neq \mathrm{ref}}^{T}\! \operatorname{WRS}\left(\tilde{S}_{l}, S_{\theta_{p}, l}\right)\!\Bigg)^{2}\!.
\end{aligned}
\end{equation}

\subsection{Uncertainty Weighting}
\label{sec-3.6}
The existing models are typically optimized with respect to the generated reports in a batch, where each report and its sentences are equally contributed to the losses. However, as the radiology reports hold similar form and the corresponding sentences describe the same region, it is greatly potential to combine multiple objective losses for each report or sentences and perform a weighted sum of multiple reports in one batch/sentences in one report. Since this weight selection is important but sensitive, our report-level and sentence-level uncertainties can be treated as Rep-weighting and Sen-weighting to learn how to balance this weighting optimally. Given a batch of input images $\{\mathbf{X}\}_{r=1}^{R}$ and the corresponding ground truth reports $\{\mathbf{R^{*}}\}_{r=1}^{R}$, the overall loss can be weighted as
\begin{equation}
    \begin{aligned}
&\ell(\{\mathbf{X}\}, \{\mathbf{R^{*}}\}) =\lambda_{AutoEn}\ell_{AutoEn} + \ell_{Rep},\\
&
\begin{aligned}
\ell_{Rep}\!=\!\sum_{r=1}^{R} &\underbrace{\!\Bigg(\!\!\exp \!\!\bigg[\!-\!\Big(\!\alpha \hat{\boldsymbol{\sigma}}^2_{\text{SMAS,}r}\!+\!\beta( \overline{\exp(\overline{\boldsymbol{\mu}_{\text{VIS,}r}'})\!+\!\boldsymbol{\sigma'}^2_{\text{VIS,}r}})\!\Big)\!\bigg] \!\Bigg)}_{\text{Rep-Weighting}} \!\times \\ 
& \Bigg(\sum_{l=1}^{L_{r}} \underbrace{\exp [-\gamma \hat{\boldsymbol{\sigma}}^2_{\text{SMAS-}l,r}]}_{\text{Sen-Weighting}}  \ell_{Sen}\left(S_{r, l}, S^{*}_{r, l}\right)\!\!\Bigg),
\end{aligned}
\end{aligned}
\end{equation}
where $\lambda_{AutoEn}$, $\alpha$, $\beta$ and $\gamma$ are four values to adjust $\ell_{AutoEn}$ and the uncertainty weights, $\beta\overline{(\cdot)}$ denotes the mean over all pixels, $S_{r, l}$ and  $S^{*}_{r, l}$ represent the $l^{th}$ sentence in the $r^{th}$ generated report and the corresponding ground-truth report in the batch, and $\ell_{Sen}$ calculates their cross entropy loss. This mechanism enables the model to learn an optimized set of weights for each sentence and each report. Higher uncertainty values ($\left|\hat{\boldsymbol{\sigma}}_{\text{SMAS,}r}\right|$ and $\left|\boldsymbol{\sigma'}_{\text{VIS,}r}\right|$) will decrease the contribution of $r^{th}$ report and higher $\left|\hat{\boldsymbol{\sigma}}_{\text{SMAS-}l,r}\right|$ leads to lower weighting of $l^{th}$ sentence in all $L_{r}$ sentences.

\section{Experimental Results}
\label{exp-overall}
The experiments are conducted based on PyTorch and an NVIDIA Tesla V100 32GB GPU with two public datasets: the Indiana University chest X-Ray (IU X-Ray) and a public COVID-19
CT Report dataset (COV-CTR). IU X-Ray consists of $7,470$ chest X-ray images with $3,955$ reports. We implement the same way as \cite{Wang_2021_CVPR} to randomly select 10\% reports for testing. The COV-CTR dataset consists $728$ images ($349$ for COVID-19 and $379$ for Non-COVID), which are collected from published papers and their corresponding paired Chinese reports by Li et al. \cite{li2020auxiliary}. We also split the entire data into train/val/test as 7:1:2, following \cite{Wang_2021_CVPR,li2020auxiliary}.

Targeting at medical domain, we adopt a 200-dimensional word embedding BioWordVec \cite{zhang2019biowordvec} trained by fastText \cite{bojanowski2017enriching} and 700-dimensional sentence embedding BioSentVec \cite{chen2019biosentvec} trained by sent2vec \cite{pagliardini2017unsupervised} as pre-trained vectors to calculate SMAS.

For a fair comparison with other transformer-based methods, we set the transformer as three blocks, $8$ heads and $512$ hidden units as in \cite{report07}. The learning rate of visual extractor is set by $5e^{-5}$, while the learning rate for the remaining parameters is set by $1e^{-4}$. The hyper-parameters $\alpha, \beta, \gamma, \lambda$ are all set to $1$ according to the best performance on the validation set. Beam Search is utilized to generate final reports with beam size $3$. All the models are trained for $50$ epochs and the best results are reported based on the validation set.
In order to accelerate the training process, we set $T=4$ for training and $T=10$ for testing. In both cases dropout is applied with $p=0.5$.
All the models are evaluated by BLEU \cite{papineni2002bleu}, METEOR \cite{banerjee2005meteor} and ROUGE-L \cite{lin2004rouge}.
\subsection{Performance Comparison}
\subsubsection{Automatic Evaluation}
\begin{figure*}[t]
\includegraphics[width=2\columnwidth]{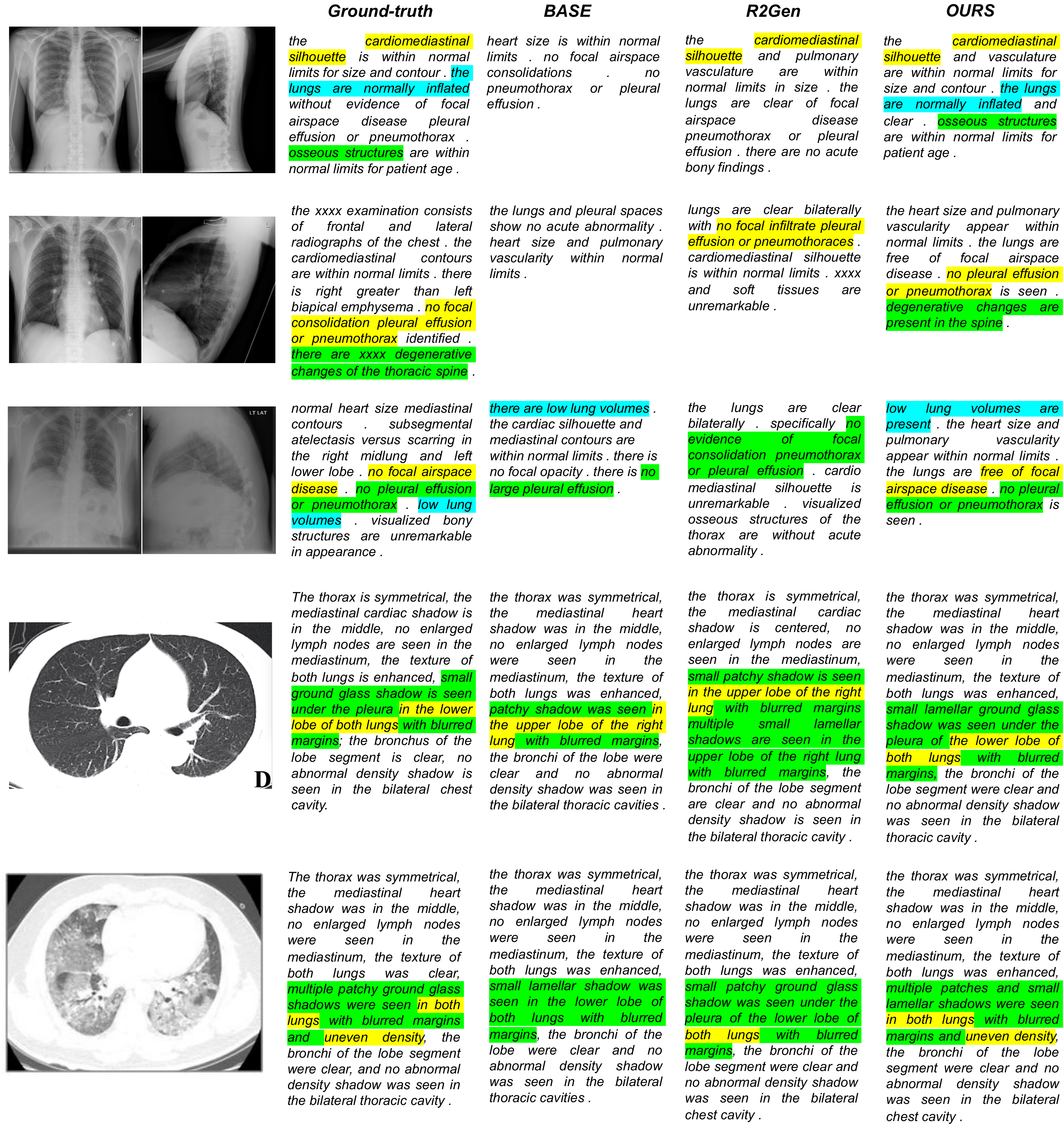}
\caption{Illustration of reports generated by BASE model, M-T Transformer and OURS. The left two columns show the input front and lateral chest X-ray images from IU X-Ray dataset (row 1, 2, 3) and CT images from COV-CTR (row 4, 5), followed by their ground-truth reports. Different colors indicate different key information.}
\label{compare}
\end{figure*}

Table \ref{tab:comparison} shows the quantitative results of our proposed methods. First, we illustrate the effectiveness of our AutoEncoder and two weighting methods separately. "BASE" represents the vanilla transformer architecture for image captioning. It is observed that the "AutoEncoder" enhances the performance on all the metrics for the highest $3.3\%$, which suggests that the AutoEncoder brings stronger ability for feature extraction. Note that applying "Sen-Weighting" and "Rep-Weighting" provides a large improvement. This indicates that using uncertainty to weight the loss can effectively optimize the process for reports/sentences generation. 
"OURS" represents the combination of all of the aforementioned strategies. It outperforms the state-of-the-art approaches, such as R2Gen \cite{report07}, CMN \cite{report10}, Self-Boosting \cite{report09} and PPKED \cite{report08}. 
The qualitative results are shown in Fig. \ref{compare}, which contains three cases from IU X-Ray dataset, including one normal radiology image (first row) and two images with some abnormalities (second \& third rows). In the first case, our model has the best ability to describe each region and generates more accurate and professional expressions, such as \textit{"cardiomediastinal silhouette"}, \textit{"osseous structures"} and \textit{"normally inflated"}. The other two images contain abnormalities about \textit{"degenerative spine"} and \textit{"low lung volumes"}, in which our model is able to capture all the symptoms while the competing methods fail to do so. The last two rows in Fig. \ref{compare} indicate two abnormal cases from COV-CTR dataset. It can also be observed that our method generates more accurate reports through finding the key information, such as \textit{"the lower lobe of both lungs"} and \textit{"uneven density"}.
\subsubsection{Human Evaluation}
To better assess the quality of our method, we also conduct a more valid and reliable expert evaluation. Three experienced radiologists were invited to evaluate the reports from three quality dimensions: Accuracy, Informativeness and Readability. They are required to give an overall 5-scale grades (5: Accept, 4: weakly accept, 3: borderline, 2: weakly reject, 1: Reject) based on the clinical acceptance. We randomly select 50 images from the testing set from IU X-Ray dataset and provide their corresponding predicted reports from our model and R2Gen respectively. The average score of our method is $3.61$ and R2Gen achieves $3.56$. According to the most experienced doctor's evaluation, $24\%$ portions of our reports are better than reports from R2Gen and $60\%$ portions are the same. The evaluation results indicate that our method can generate much more acceptable reports in real clinical practice.
\begin{table*}[htbp]
  \centering
  \caption{Comparison results. $\dagger$ refers to results from the original papers. $\ddagger$ refers to our reproduction results from officially released codes.}
  \scalebox{1.1}{
    \begin{tabular}{c|l|cccccc}
    \toprule
    \multicolumn{1}{c|}{\multirow{2}[4]{*}{Dataset}} & \multicolumn{1}{c|}{\multirow{2}[4]{*}{Model}} & \multicolumn{6}{c}{Evaluation Metrics} \\
\cmidrule{3-8}      &    & \multicolumn{1}{c}{BLEU-1} & \multicolumn{1}{c}{BLEU-2} & \multicolumn{1}{c}{BLEU-3} & \multicolumn{1}{c}{BLEU-4} & \multicolumn{1}{c}{METEOR} & \multicolumn{1}{c}{ROUGE-L} \\
    \midrule
    \multicolumn{1}{c|}{\multirow{10}[4]{*}{IU-Xray}} & BASE  & 0.446  & 0.280  & 0.253  & 0.159  & 0.182  & 0.359  \\
    & + AutoEncoder & 0.465  & 0.320  & 0.235  & 0.175  & 0.192  & 0.392  \\
    & + AutoEncoder + VISVar &  0.469 & 0.327 & 0.250 & 0.198 & 0.202 & 0.399  \\
    & + AutoEncoder + Sen-Weighting& 0.485  & 0.339  & 0.253  & 0.192  & 0.189  & 0.395  \\
    & + AutoEncoder + Rep-Weighting & \textbf{0.494}  & \textbf{0.351}  & \textbf{0.272}  & \textbf{0.219}  & \textbf{0.217}  & \textbf{0.414}  \\
    \cmidrule{2-8}
    \cmidrule{2-8}
    & $\text{COATT \cite{COATT}}^{\dagger}$ & 0.455 & 0.288 & 0.205 & 0.154 & - & 0.369 \\
    & $\text{HRGR-Agent \cite{HRGR-Agent}}^{\dagger}$ & 0.438 & 0.298 & 0.208 & 0.151 & - & 0.322 \\
    & $\text{CMAS-RL \cite{CMAS-RL}}^{\dagger}$ & 0.464 & 0.301 & 0.210  & 0.154 & - & 0.362 \\
    & $\text{R2Gen \cite{report07}}^{\dagger}$ & 0.470  & 0.304  & 0.219  & 0.165  & 0.187  & 0.371  \\
    & $\text{CMN \cite{report10}}^{\dagger}$ & 0.475 & 0.309 & 0.222 & 0.170 & 0.191 & 0.375   \\
    & $\text{Self-Boosting \cite{report09}}^{\dagger}$ & 0.487 & 0.346 & 0.270 & 0.208 & - & 0.359\\
    & $\text{PPKED \cite{report08}}^{\dagger}$ & 0.483 & 0.315 & 0.224 & 0.168 & \textbf{0.376} & 0.351 \\
    & \textbf{OURS} & \textbf{0.497}  & \textbf{0.357}  & \textbf{0.279}  & \textbf{0.225}  & 0.217  & \textbf{0.408}  \\
    \midrule
    \midrule
    \multicolumn{1}{c|}{\multirow{4}[4]{*}{COV-CTR}} & BASE  & 0.710  & 0.628  & 0.561  & 0.505  & 0.427  & 0.689  \\
    & $\text{COATT \cite{COATT}}^{\dagger}$ & 0.709 & 0.645 & 0.603 & 0.552 & - & 0.718  \\
    & $\text{Vision-BERT \cite{devlin2019bert}}^{\dagger}$ & 0.710  & 0.653  & 0.606 & 0.558  & -  & \textbf{0.747}  \\
    & $\text{ASGK \cite{li2020auxiliary}}^{\dagger}$ & 0.712  & 0.659  & 0.611 & \textbf{0.570}  & -  & 0.746  \\
    & $\text{R2Gen \cite{report07}}^{\ddagger}$ & 0.731  & 0.647  & 0.579  & 0.521  & 0.426  & 0.699  \\
    & \textbf{OURS} & \textbf{0.753}  & \textbf{0.680}  & \textbf{0.620}  & 0.569  & \textbf{0.437}  & 0.730  \\
    \bottomrule
    \end{tabular}}%
  \label{tab:comparison}%
\end{table*}%

\begin{figure}[t]
\centerline{\includegraphics[width=1\columnwidth]{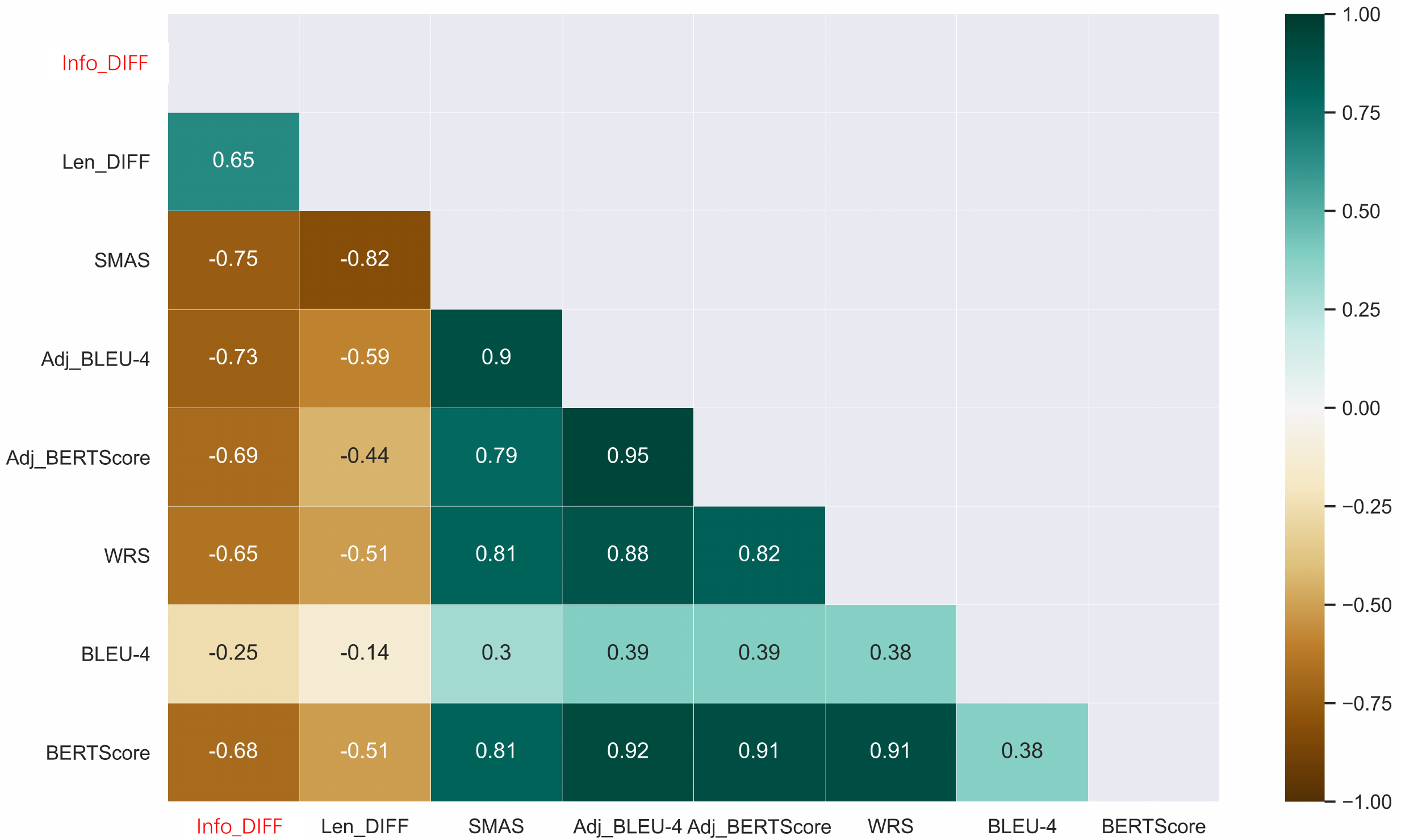}}
\caption{Heatmap of correlation among different similarity measurements. "Info\_DIFF" means key information difference of two diagnostic reports annotated by experts. Greater negative correlation between other scores and "Info\_DIFF" represents better similarity measurement.}
\label{heatmap}
\end{figure}

\subsection{Report-level \& Sentence-level Uncertainty} \label{section: Report-level & Sentence-level Uncertainty}

\subsubsection{SMAS}\label{smas-exp} Since the similarities of reports should be largely determined by the key information such as diseases and organs, we randomly select 60 reports and manually annotate their corresponding diseases and organs. For example, the sentence "heart size mildly enlarged" will be labeled by its organ \textit{"Heart"} and abnormality \textit{"Large"}. 
Then we randomly sample 2500 pairs of reports out of the 60 samples with replacement. To illustrate that our SMAS can better capture the similarities among different reports, we calculate the similarities on these pairs using different similarity scores and obtain a heatmap in Fig. \ref{heatmap} to show their correlations. "Info\_DIFF" and "Len\_DIFF" represent the absolute value of the differences on the number of labels and sentences bewteen two reports. "WRS", "BLEU-4" and "BERTScore" calculate the similarity scores by treating full report as a single sentence. "Adj\_BLEU-4" and "Adj\_BERTScore" represent applying our $\operatorname{MATCH}(\cdot)$ function to BLEU-4 and BERTScore.
First, we note that the correlation between "Info\_DIFF" and "Len\_DIFF" is $0.65$, which means that we can use the inconsistency in the sentence number to roughly determine the inconsistency between reports. This is useful as extracting key information from reports manually is labor-some.
Moreover, we find that there is a significantly higher negative correlation between the "Adj\_BLEU-4" vs. the "BLEU-4" and "Info\_Diff" ($-0.73$ vs. $-0.25$), which means the former gives a more faithful evaluation on the generated reports' relationships by utilizing the key information to measure similarity. Such improvements is solely achieved through our novel $\operatorname{MATCH}(\cdot)$ algorithm.
Lastly, note that our "SMAS" has the greatest negative correlation to "Info\_DIFF", proving the strong ability to measure report similarities mostly based on key information. 
\subsubsection{Report-level Uncertainty}
\label{exp-report-level-uncertainty}
\begin{figure*}[t]
  \centering
  \caption{Illustration of difference between BLEU-4 and report-level uncertainty.}
  \label{fig:report-level}
  \includegraphics[width=2\columnwidth]{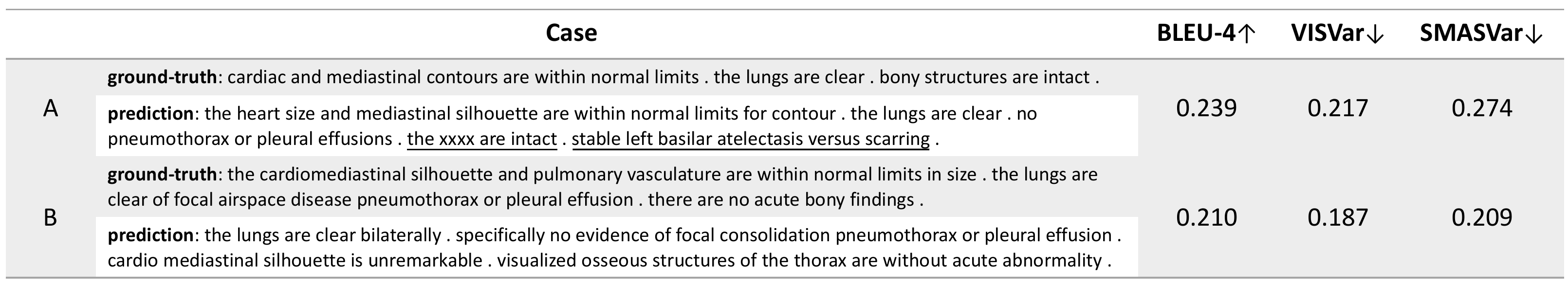}
\end{figure*}

\begin{figure*}[t]
  \centering
  \includegraphics[width=2\columnwidth]{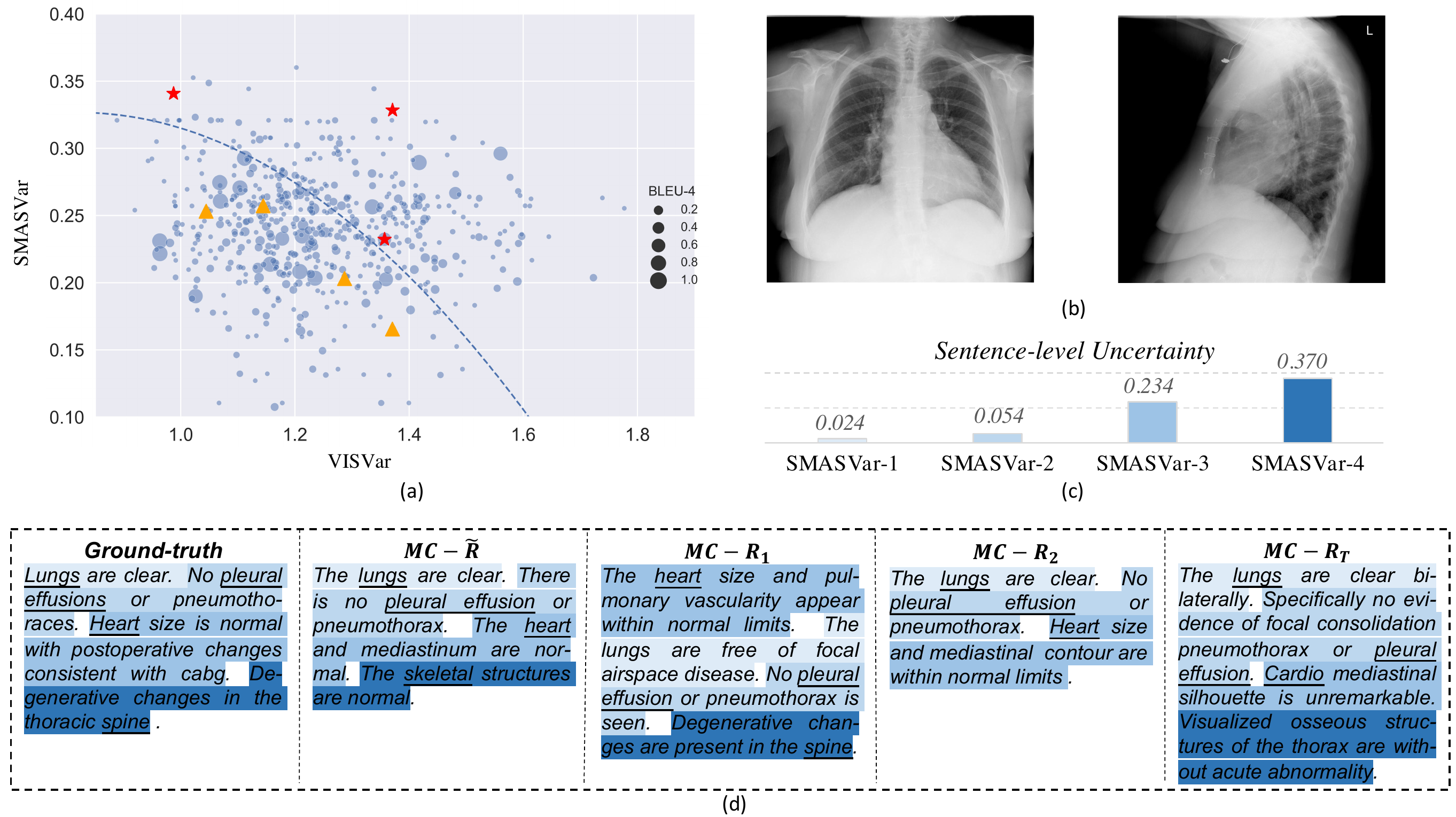}
  \caption{Illustration of the report-level and sentence-level uncertainties. a) The relationship among \textbf{SMASVar}, \textbf{VISVar} and BLEU-4. The threshold curve is adjustable to find highly uncertain cases for risk assessment. Red stars / Orange triangles show examples that real diagnosis variance is 1 / 0 by human evaluation. b) Front and lateral chest X-ray images. c) Sentence-level uncertainty. d) Ground-truth, reference report $MC-\widetilde{R}$, and several MC outputs. Different colors show different uncertainty values of corresponding matched sentences. }
  \label{fig:experiment}
\end{figure*}

The report-level uncertainty is reflected by \textbf{SMASVar} and \textbf{VISVar}. We first draw a scatter plot from the testing set in Fig. \ref{fig:experiment}(a) to show the relationship among \textbf{SMASVar}, \textbf{VISVar} and BLEU-4 score. We set a threshold curve to find out the cases with extreme high uncertainty. 
The threshold can be modified based on doctors' risk assessment in clinical practice. This report-level uncertainty serves as an extra important evaluation to assess the reliability or confidence of the generated reports. Fig. \ref{fig:report-level} shows two cases where A has a higher BLEU-4 score than B. However, the prediction of case A contains two redundant uncertain sentences and their \textbf{VISVar} and \textbf{SMASVar} show higher value than case B, which indicates the information contained in A is less confident than those in B. The superiorities of \textbf{VISVar} and \textbf{SMASVar} over SMAS are further discussed in Appendix\ref{Report-level Uncertainty Samples}. It is also noted that if most generated sentences in one report are certain while only one is uncertain, \textbf{SMASVar} will still be lower. The proposed sentence-level uncertainty $\textbf{SMASVar-}\boldsymbol{l}$ can tackle this problem and measure each sentence's uncertainty concisely. To validate our report-level uncertainty in real clinical settings, we compare it with real diagnosis variance of human evaluation. The variance of the scores from three experienced doctors could represent the real clinical uncertainty. As shown in Fig. \ref{fig:experiment}(a), red stars indicate diagnosis reports that obtain the same score from all three doctors, i.e., variance = 0, and the orange triangles represent cases that three doctors have strong disagreement on, i.e., variance = 1. It is noted that cases with higher (lower) doctor variance also have larger (smaller) \textbf{VISVar} and \textbf{SMASVar}. Therefore, our report-level uncertainty is comparable to real doctors' diagnosis variance and setting a proper threshold can help doctor make further decision on whether to follow up the automatic generated reports.

\subsubsection{Sentence-level Uncertainty} Fig. \ref{fig:experiment}(b)-(d) show the meanings of sentence-level uncertainty, where one case is presented, along with its several predictions $MC-R_1$ to $MC-R_T$ under $T$ times MC dropout sampling and the reference report $MC-\widetilde{R}$ among them. Different colors represent the uncertainty values of the corresponding sentences. 
Our proposed reference report covers every information in the ground truth including lungs, pleural effusion, heart and spine with different uncertainties. One can easily observe that the sentence describing spine has the highest uncertainty, which can be explained through the uncertain predicted sentence on spine from $MC-R_1$ to $MC-R_T$. This suggests that the model cannot make sure whether the spine is normal or not. Such an uncertainty should be warned by the report generation system so that doctors can make further decisions based on it.
\section{Conclusion}
\label{conclusion}
In this paper, we develop a novel method to explicitly quantify the model uncertainties in both the visual and textual domain for the task of radiology report generation. Such multi-modal uncertainty measurements can sufficiently capture the model confidence scores at both the report and sentence level, which can be further leveraged to weight the losses to achieve more comprehensive model optimization. Our experimental results have demonstrated that our proposed method outperforms all existing approaches. Most importantly, extensive experiments have also validated its ability to capture the uncertainty in radiology report generation systems.

At present, our model only builds relationships between the visual and textual uncertainties at the report-level. We intend to explicitly map the sentence-level uncertainty with the region-specific uncertainty in images, which will improve the model performance significantly, lead to better model interpretability, and provide better guidance in clinical practice. Future works also include introducing more prior knowledge from medical domain to enhance the reliability of the generation models.

\appendix
\label{Report-level Uncertainty Samples}
\begin{figure*}[htbp]
  \centering
  \includegraphics[width=2\columnwidth]{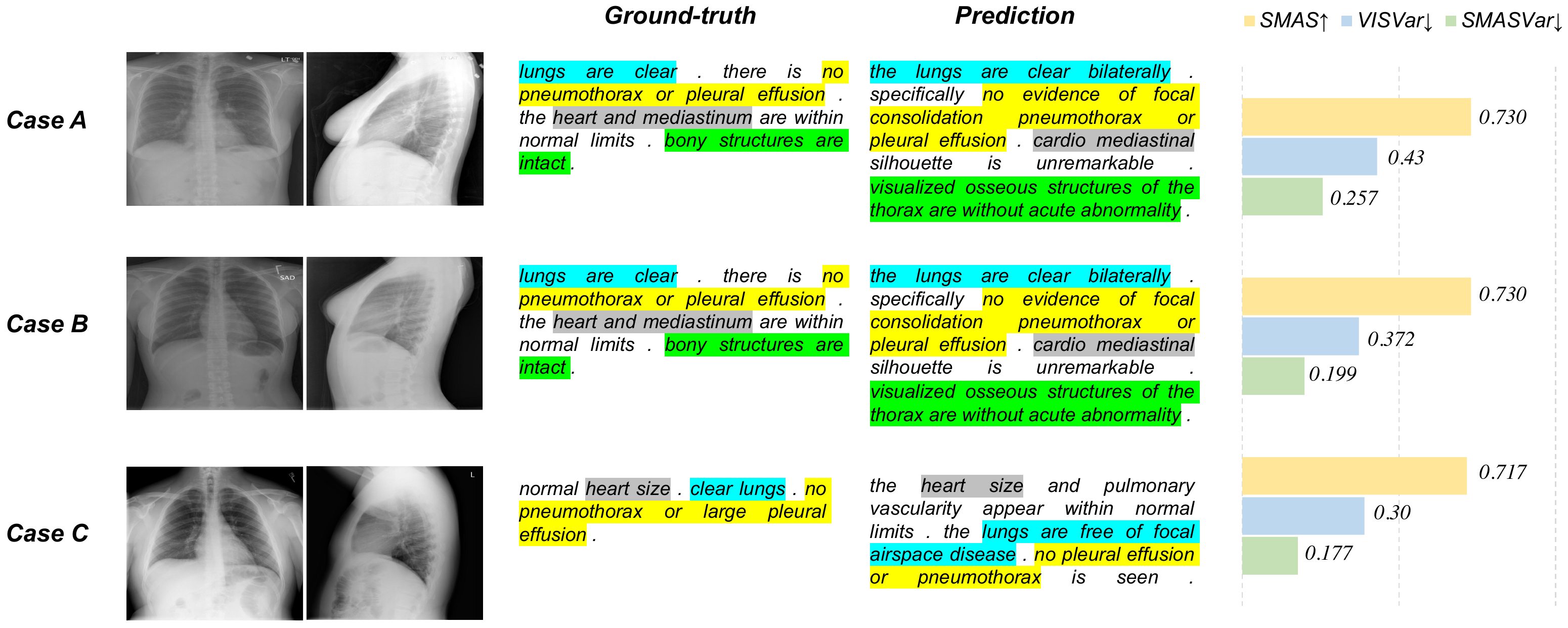}
  \caption{Samples of report-level uncertainties.  Lower \textbf{VISVar} and \textbf{SMASVar} and higher SMAS indicate better performance.}
  \label{fig: appendix-report-level-uncertainty-samples}
\end{figure*}
\begin{figure*}[htbp]
  \centering
  \includegraphics[width=2\columnwidth]{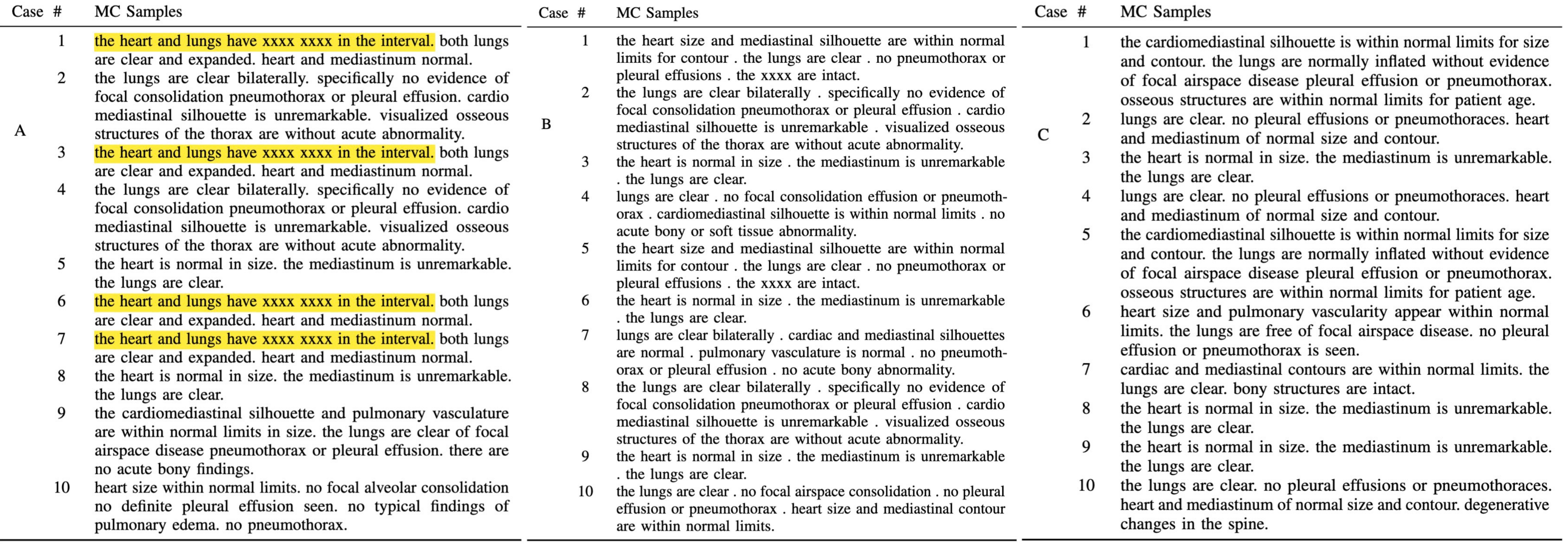}
  \caption{Generated $MC$ predictions from case A, B, C under $T=10$ times $MC$ dropout sampling.}
  \label{appendix-mc}
\end{figure*}
We illustrate how \textbf{SMASVar} and \textbf{VISVar} serve as the confidence evaluation methods. Fig. \ref{fig: appendix-report-level-uncertainty-samples} shows three cases A, B and C, all of which are predicted correctly by the model. They have similar SMAS values as SMAS evaluates the results only from the similarity of ground-truths and predictions. However, due to the different input X-ray images, the model holds different confidence in their predictions, which can be captured by our methods in 
Fig. \ref{appendix-mc}. With $T=10$ times MC dropout, the generated $10$ predictions from the input images of case A are much uncertain while the $10$ predictions are similar from B and C. Though A and B have the same SMAS, the MC-\#1/\#3/\#6/\#7 of case A include wrong information "\textit{the heart and lungs have xxxx xxxx in the interval}", which makes the final prediction not as reliable as case B. Therefore, the \textbf{SMASVar} and \textbf{VISVar} of B are rather smaller than A, which means the model is more confident and certain in B's X-ray images. 



\end{document}